\documentclass[10pt,twocolumn,letterpaper]{article}

\usepackage{cvpr}
\usepackage{times}
\usepackage{epsfig}
\usepackage{graphicx}
\usepackage{amsmath}
\usepackage{amssymb}
\usepackage{xcolor}
\usepackage{subfig}
\usepackage{algorithm}
\usepackage{algorithmicx}
\usepackage{algpseudocode} 
\usepackage{threeparttable}
\usepackage{multirow}
\usepackage{multicol}
\usepackage{booktabs}
\usepackage{subfig}
\usepackage{rotating}
\cvprfinalcopy %

\def\cvprPaperID{****} %

\ifcvprfinal\fi
\begin{document}

\title{Pruning from Scratch}

\author{Yulong Wang$^1$\thanks{Work done while an intern at Ant Financial.}, Xiaolu Zhang$^2$, Lingxi Xie$^3$, Jun Zhou$^2$,\\ Hang Su$^1$, Bo Zhang$^1$, Xiaolin Hu$^1\thanks{Corresponding author}$\\
	$^1$Tsinghua University \quad $^2$Ant Financial \quad $^3$Huawei Noah's Ark Lab\\
	{\tt\small wang-yl15@mails.tsinghua.edu.cn, \quad \{yueyin.zxl,jun.zhoujin\}@antfin.com \quad 198808xc@gmail.com}\\
	{\tt\small \{suhangss,dcszb,xlhu\}@mail.tsinghua.edu.cn}
}

\maketitle
\thispagestyle{empty}

\begin{abstract}
	Network pruning is an important research field aiming at reducing computational costs of neural networks. Conventional approaches follow a fixed paradigm which first trains a large and redundant network, and then determines which units (\textit{e.g.}, channels) are less important and thus can be removed. In this work, we find that pre-training an over-parameterized model is not necessary for obtaining the target pruned structure. In fact, a fully-trained over-parameterized model will reduce the search space for the pruned structure. We empirically show that more diverse pruned structures can be directly pruned from randomly initialized weights, including potential models with better performance. Therefore, we propose a novel network pruning pipeline which allows \textit{pruning from scratch}. In the experiments for compressing classification models on CIFAR10 and ImageNet datasets, our approach not only greatly reduces the pre-training burden of traditional pruning methods, but also achieves similar or even higher accuracy under the same computation budgets. Our results facilitate the community to rethink the effectiveness of existing techniques used for network pruning.
	
\end{abstract}

\section{Introduction}
\begin{figure}
	\centering
	\includegraphics[width=0.98\columnwidth]{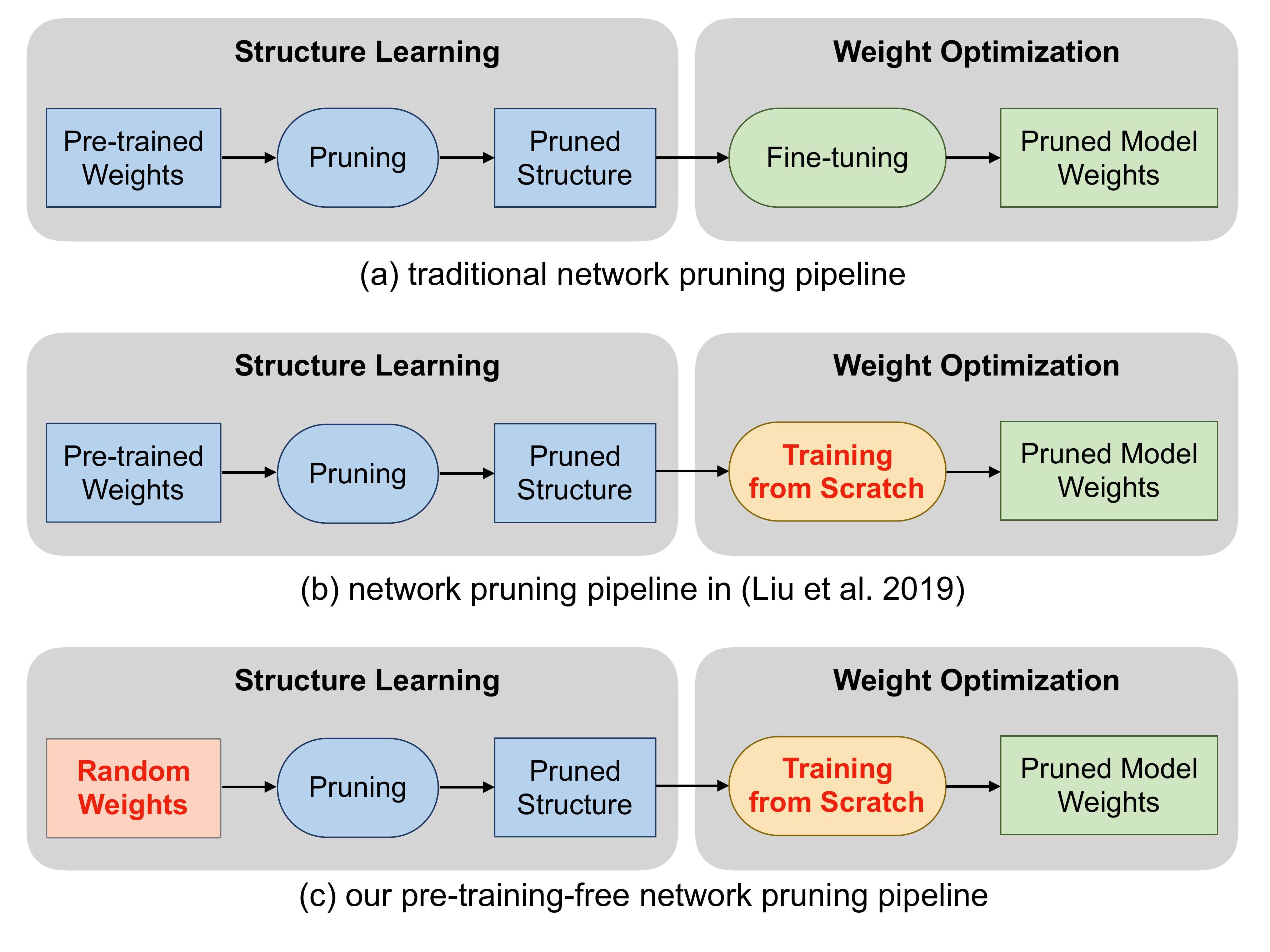}
	\caption{
		Network pruning pipelines.
		(a) Traditional network pruning needs pre-trained weights and certain pruning strategy for pruned structure learning, and fine-tuning on full model weights.
		(b) Recent work~\cite{liu2018rethinking} shows that the pruned model can be trained from scratch without fine-tuning to reach comparable performance.
		However, the pruned model structure still needs to be obtained by traditional pruning strategies.
		(c) We empirically show that the pruned model structure can be directly learned from randomly initialized weights without the loss of performance.
	}
	\label{fig:overview}
\end{figure}

As deep neural networks are widely deployed in mobile devices,
there has been an increasing demand for reducing model size and run-time latency.
Network pruning~\cite{han2015deep,channelpruning,slimming} techniques are proposed to achieve model compression and inference acceleration
by removing redundant structures and parameters. In addition to the early non-structured pruning methods~\cite{lecun1990optimal,han2015deep},
the structured pruning method represented by channel pruning~\cite{l1norm,luo2017thinet,channelpruning,slimming}
has been widely adopted in recent years because of its easy deployment on general-purpose GPUs.
The traditional network pruning methods adopt a three-stage pipeline,
namely pre-training, pruning, and fine-tuning~\cite{liu2018rethinking}, as shown in Figure~\ref{fig:overview}(a).
The pre-training and pruning steps can also be performed alternately with multiple cycles~\cite{softfilter}.
However, recent study~\cite{liu2018rethinking} has shown that the pruned model can be trained from scratch to achieve a comparable prediction performance without the need to fine-tune the inherited weights from the full model (as shown in Figure~\ref{fig:overview}(b)).
This observation implies that the pruned architecture is more important for the pruned model performance.
Specifically, in the channel pruning methods, more attention should be paid to searching the channel number configurations of each layer.

Although it has been confirmed that the \textit{weights} of the pruned model do not need to be fine-tuned from the pre-trained weights, the \textit{structure} of the pruned model still needs to be learned and extracted from a well-trained model according to different criteria. 
This step usually involves cumbersome and time-consuming weights optimization process. Then we naturally ask a question: Is it necessary for learning the pruned model structure from pre-trained weights? 

In this paper, we explored this question through extensive experiments and found that the answer is quite surprising: an effective pruned structure does not have to be learned from pre-trained weights. We empirically show that the pruned structures discovered from pre-trained weights tend to be homogeneous, which limits the possibility of searching for better structure. 
In fact, more diverse and effective pruned structures can be discovered by directly pruning from randomly initialized weights, including potential models with better performance. 

Based on the above observations, we propose a novel network pruning pipeline that a pruned network structure can be directly learned from the randomly initialized weights (as shown in Figure~\ref{fig:overview}(c)).
Specifically, we utilize a similar technique in Network Slimming~\cite{slimming} to learn the channel importance
by associating scalar gate values with each layer. The channel importance is optimized to improve the model performance
under the sparsity regularization. What is different from previous works is that we do not update the random weights
during this process. After finishing the learning of channel importance, we utilize a
simple binary search strategy to determine the channel number configurations of the pruned model given resource constraints (\textit{e.g.}, FLOPS). Since we do not need to update the model weights during optimization,
we can discover the pruned structure at an extremely fast speed. Extensive experiments on CIFAR10~\cite{krizhevsky2009learning} and ImageNet~\cite{russakovsky2015imagenet} show that our method yields at least $10\times$ and $100\times$ searching speedup while achieving comparable or even better model accuracy than traditional pruning methods using complicated strategies.
Our method can free researchers from the time-consuming training process
and provide competitive pruning results in future work.

\section{Related Work}
Network pruning techniques
aim to achieve the inference acceleration of deep neural networks by removing the redundant parameters and structures in the model.
Early works~\cite{lecun1990optimal,han2015deep,han2015learning} proposed to remove individual weight values, resulting in
non-structured sparsity in the network. The runtime acceleration cannot be easily achieved on a general-purpose GPU,
otherwise with a custom inference engine~\cite{han2016eie}.
Recent works focus more on the development of structured model pruning~\cite{l1norm,channelpruning,slimming},
especially pruning weight channels. $\ell_1$-norm based criterion~\cite{l1norm} prunes model according to the $\ell_1$-norm of weight channels.
Channel Pruning~\cite{channelpruning} learns to obtain sparse weights by minimizing local layer output reconstruction error.
Network Slimming~\cite{slimming} uses LASSO regularization to learn the importance of all channels and prunes the model based on a global threshold.
Automatic Model Compression (AMC)~\cite{he2018amc} explores the pruning strategy by automatically learning the compression ratio of each layer through reinforcement learning (RL).
Pruned models often require further fine-tuning to achieve higher prediction performance.
However, recent works~\cite{liu2018rethinking,lottery} have challenged this paradigm and
show that the compressed model can be trained from scratch to achieve comparable performance without relying on the fine-tuning process.

Recently, neural architecture search (NAS) provides another perspective on the discovery of the compressed model structure.
Recent works~\cite{liu2018darts,cai2018proxylessnas} follow the top-down pruning process by
trimming out a small network from a supernet. The one-shot architecture search methods
~\cite{brock2017smash,bender2018understanding} further develop this idea and conduct architecture search only once after
learning the importance of internal cell connections. However, these methods require a large amount of training time to search for an efficient structure.

\section{Rethinking Pruning with Pre-Training}

\begin{figure*}[ht]
	\centering
	\includegraphics[width=\textwidth]{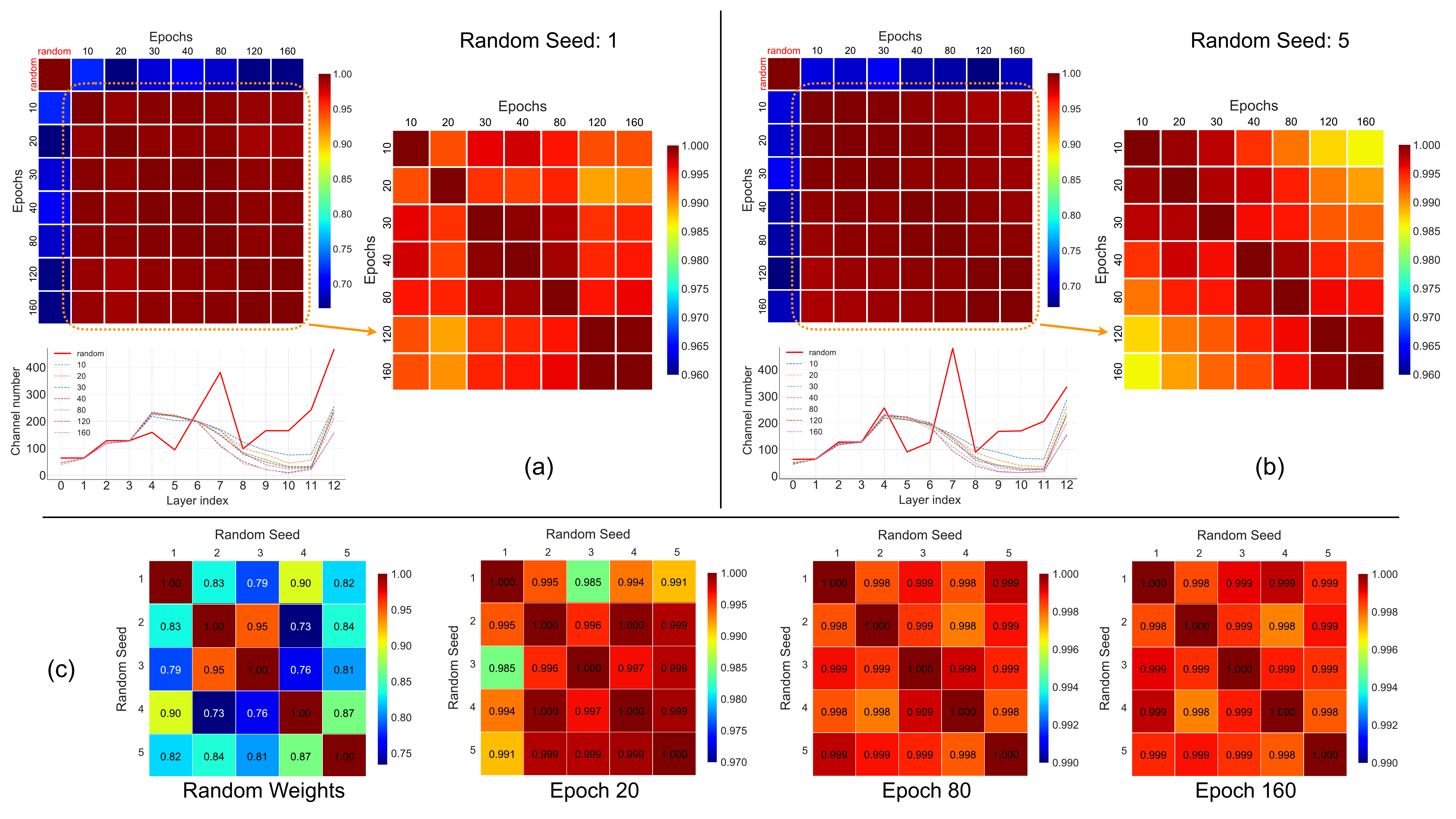}
	\caption{Exploring the effect of pre-trained weights on the pruned structures. All the pruned models are required to reduce 50\% FLOPS of the original VGG16 on CIFAR10 dataset. (a) (Top-left) We display the correlation coefficient matrix of the pruned models directly learned from randomly initialized weights (``\textbf{random}'') and other pruned models based on different checkpoints during pre-training (``\textbf{epochs}''). (Right) We display the correlation coefficient matrix of pruned structures from pre-trained weights on a finer scale. (Bottom-left) We show the channel numbers of each layer of different pruned structures. Red line denotes the structure from random weights. (b) Similar results from the experiment with a different random seed. (c) We display correlation coefficient matrices of all the pruned structures from five different random seeds. We mark the names of initialized weights used to get pruned structures below.}
	\label{fig:final_compare}
\end{figure*}

Network pruning aims to reduce the redundant parameters or structures in an over-parameterized model to obtain an efficient pruned network. Representative network pruning methods~\cite{slimming,gordon2018morphnet} utilize channel importance to evaluate whether a specific weight channel should be reserved. Specifically, given a pre-trained model, a set of channel gates are associated with each layer to learn the channel importance. The channel importance values are optimized with $\ell_1$-norm based sparsity regularization.  Then with the learned channel importance values,
a global threshold is set to determine which channels are preserved given a predefined resource constraint. The final pruned model weights can either be fine-tuned from the original full model weights or re-trained from scratch. The overall pipeline is depicted in Figure~\ref{fig:overview}(a) and (b). 

In what follows, we show that in the common pipeline of network pruning, the role of pre-training is quite different from what we used to think. Based on this observation, we present a new pipeline which allows pruning networks from scratch, \textit{i.e.}, randomly initialized weights, in the next section.

\subsection{Effects of Pre-Training on Pruning}

The traditional pruning pipeline seems to default to a network that must be fully trained before it can be used for pruning.
Here we will empirically explore the effect of the pre-trained weights on the final pruned structure. Specifically, we save the checkpoints after different training epochs when we train the baseline network. Then we utilize the weights of different checkpoints as the network initialization weights, and learn the channel importance of each layer 
by adopting the pipeline described above. We want to explore whether the pre-trained weights at different training stages have a crucial impact on the final pruned structure learning.

\subsection{Pruned Structure Similarity} 
First, we compare the structure similarity between different pruned models. For each pruned model, we calculate the pruning ratio of each layer, \textit{i.e.}, the number of remaining channels divided by the number of original channels. The vector formed by concatenating the pruning ratios of all layers is then considered to be the feature representation of the pruned structure. Then we calculate the correlation coefficient between each of the two pruned model features as the similarity of their structures. In order to ensure the validity, we randomly selected five sets of random seeds for experiments on CIFAR10 dataset with VGG16~\cite{simonyan2014very} network. We include more visualization results of ResNet20 and ResNet56~\cite{he2016deep} in the supplementary material. 

Figure~\ref{fig:final_compare} shows the correlation coefficient matrices for all pruned models. From this figure, we can observe three phenomena. First, the pruned structures learned from random weights are not similar to all the network structures obtained from pre-trained weights (see top-left figures in Figure~\ref{fig:final_compare}(a)(b)). Second, the pruned model structures learned directly from random weights are more diverse with various correlation coefficients. Also, after only ten epochs of weights update in the pre-training stage, the resulting pruned network structures become almost homogeneous. (see Figure~\ref{fig:final_compare}(c)). Third, the pruned structures based on the checkpoints from near epochs are more similar with high correlation coefficients in the same experiment run (see right figures in Figure~\ref{fig:final_compare}(a)(b)). 

The structure similarity results indicate that the potential pruned structure space is progressively reduced during the weights update in the pre-training phase, which may limit the potential performance accordingly. On the other hand, the randomly initialized weights allow the pruning algorithm to explore more diverse pruned structures.

\subsection{Performance of Pruned Structures} 

We further train each pruned structure from scratch to compare the final accuracy. Table~\ref{tab:final_compare_acc} summarizes the prediction accuracy of all pruned structures on the CIFAR10 test set. It can be observed that the pruned models obtained from the random weights can always achieve comparable performance with the pruned structures based on the pre-trained weights. Also, in some cases (such as ResNet20), the pruned structures directly learned from random weights achieves even higher prediction accuracy. These results demonstrate that not only the pruned structures learned directly from random weights are more diverse, but also that these structures are valid and can be trained to reach competitive performance.

\begin{table}[!t]
	\centering
	\caption{Pruned model accuracy (\%) on the CIFAR10 test set. All models are trained from scratch based on the training scheme in~\cite{liu2018rethinking}. We report the average accuracy across five runs. ``Rand'' stands for pruned structures from random weights. ``RN'' stands for ResNet.}
	\vspace{-0.2cm}
	\scalebox{0.9}{ %
		\renewcommand\tabcolsep{2.5pt} %
		\begin{threeparttable}
			\begin{tabular}{lcccccccc}
				\toprule
				\multirow{2}[4]{*}{Model} & \multirow{2}[4]{*}{Rand} & \multicolumn{7}{c}{Pre-training Epochs} \\
				\cmidrule{3-9}          &       & 10    & 20    & 30    & 40    & 80    & 120   & 160 \\
				\midrule
				VGG16 & 93.68 & 93.60 & 93.83 & 93.71 & 93.69 & 93.64 & 93.69 & 93.58 \\
				RN20 & 90.57 & 90.48 & 90.50 & 90.49 & 90.33 & 90.42 & 90.34 & 90.23 \\
				RN56 & 92.95 & 92.96 & 92.90  & 92.98 & 93.04 & 93.03 & 92.99 & 93.05 \\
				\bottomrule
			\end{tabular}%
		\end{threeparttable}
	}
	\label{tab:final_compare_acc}%
\end{table}%

The pruned model accuracy results also demonstrate that the pruned structures based on pre-trained weights have little advantages in the final prediction performance. Considering that the pre-training phase often requires a cumbersome and time-consuming computation process, we think that network pruning can directly start from randomly initialized weights.

\section{Our Solution: Pruning from Scratch}
Based on the above analysis, we propose a new pipeline named \textit{pruning from scratch}. Different from existing ones, it enables researchers to obtain pruned structure directly from randomly initialized weights.

Specifically, we denote a deep neural network as $f(x;\boldsymbol{W},\alpha)$, where $x$ is an input sample, $\boldsymbol{W}$ is
all trainable parameters, and $\alpha$ is the model structure. In general, $\alpha$ includes operator types,
data flow topology, and layer hyper-parameters as modeled in NAS research.
In the network pruning, we mainly focus on the micro-level layer settings, especially the channel
number of each layer in the channel pruning strategies.

To efficiently learn the channel importance for each layer, a set of scalar gate values $\boldsymbol{\lambda}_j$ are associated with
the $j$-th layer along the channel dimension. The gate values are multiplied onto the layer's output to perform channel-wise modulation.
Therefore, a near-zero gate value will suppress the corresponding channel output, resulting in a pruning effect. We denote
the scalar gate values across all the $K$ layers as $\Lambda = \{\boldsymbol{\lambda}_1, \boldsymbol{\lambda}_2, \cdots, \boldsymbol{\lambda}_K\}$.
The optimization objective for $\Lambda$ is
\begin{align}
\min_\Lambda & \quad \sum_i^N \mathcal{L}(f(x_i;\boldsymbol{W},\Lambda), y_i) + \gamma\sum_j^K |\boldsymbol{\lambda}_j|_1 \cr
s.t. & \quad \boldsymbol{0} \preceq \boldsymbol{\lambda}_j \preceq \boldsymbol{1},\quad \forall j = 1,2,\cdots,K,
\end{align}
where $y_i$ is the corresponding label, $\mathcal{L}$ is cross-entropy loss function, $\gamma$ is a balance factor. Here, the difference from previous works is two-fold. First, we do not update the weights during channel importance learning; Second, we use randomly initialized weights without relying on pre-training.

Following the same approach in Network Slimming, we adopt sub-gradient descent to optimize $\Lambda$ for the non-smooth regularization term.
However, the naive $\ell_1$-norm will encourage the gates to be zeroes unconstrainedly,
which does not lead to a good pruned structure.
Different from the original formulation in Network Slimming, we use the element-wise mean of all the gates to approximate the overall sparsity ratio,
and use the square norm to push the sparsity to a predefined ratio $r$~\cite{luo2018autopruner}. Therefore, given a target sparsity ratio $r$, the regularization term is 
\begin{equation}
\Omega(\Lambda) = (\frac{\sum_j|\boldsymbol{\lambda}_j|_1}{\sum_j C_j} - r)^2, 
\end{equation}
where $C_j$ is the channel number of the $j$-th layer. Empirically, we find this improvement can obtain more reasonable pruned structure.
During the optimization, there can be multiple possible gates for pruning.
We select the final gates whose sparsity is below the target ratio $r$ while achieving the maximum validation accuracy.

After obtaining a set of optimized gate values $\Lambda^*=\{\boldsymbol{\lambda}_1^*, \boldsymbol{\lambda}_2^*,\cdots,\boldsymbol{\lambda}_n^*\}$,
we set a threshold $\tau$ to decide which channels are pruned. In the original Network Slimming method, the global pruning threshold is determined according to a predefined reduction ratio of the target structure's parameter size. However, a more practical approach is to find the pruned structure based on the FLOPS constraints of the target structure. A global threshold $\tau$ can be determined by binary search until the pruned structure satisfies the constraints.

Algorithm~\ref{algo} summarizes the searching strategy.
Notice that a model architecture generator $G(\cdot)$ is required to generate a model structure given a set of channel number configurations.
Here we only decide the channel number of each convolutional layer and do not change the original layer connection topology.

\begin{algorithm}[ht]
	\begin{algorithmic}[1]
		\Require Optimized channel gate values $\Lambda^*$, maximum FLOPS $C$, model architecture generator $G(\Lambda)$, iterations $T$, relative tolerance ratio $\epsilon$, $\tau_{\min}=0, \tau_{\max}=1$
		\Ensure Final threshold $\tau^*$, pruned model architecture $\mathcal{A^*}$
		\For{$t\gets 1$ to $T$}
		\State $\tau_t = \frac{1}{2}(\tau_{\min} + \tau_{\max})$
		\State Get pruned channel gates $\Lambda_t$ by threshold $\tau_t$
		\State Get pruned model architecture $\mathcal{A}_t = G(\Lambda_t)$
		\State $C_t=\mathtt{calculate\_FLOPS}(\mathcal{A}_t)$
		\If{$|C_t - C| / C \leq \epsilon$}
		\State $\tau^*=\tau_t,\,\, \mathcal{A^*}=\mathcal{A}_t$
		\State break
		\EndIf
		\State \textbf{if} $C_t < C$ \textbf{then} $\tau_{\min}=\tau_t$ \textbf{else} $\tau_{\max}=\tau_t$
		
		\EndFor
	\end{algorithmic}
	\caption{Searching For Pruned Structure}
	\label{algo}
\end{algorithm}

\subsection{Implementations}
\subsubsection{Channel Expansion}
The new pruning pipeline allows us to explore a larger model search space with no cost. We can change the full model size and then obtain the target pruned structure by
slimming network. The easiest way to change model capacity is to use uniform channel expansion, which uniformly enlarges or shrinks the channel numbers of all layers with a common width multiplier.
As for the networks with skip connection such as ResNet~\cite{he2016deep},
the number of final output channels of each block and the number of channels at the block input are simultaneously
expanded by the same multiplier to ensure that the tensor dimensions are the same.

\subsubsection{Budget Training}
A significant finding in~\cite{liu2018rethinking} is that a pruned network can achieve similar performance
to a full model as long as it is adequately trained for a sufficient period.
Therefore, the authors in~\cite{liu2018rethinking} proposed ``Scratch-B'' training scheme,
which trains the pruned model for the same amount of computation budget with the full model.
For example, if the pruned model saves $2\times$ FLOPS, we double the number of basic training epochs,
which amounts to a similar computation budget. Empirically, this training scheme is crucial for improving the pruned model performance.

\subsubsection{Channel Gates Location}
Following the same practice in Network Slimming~\cite{slimming}, we associate the channel gates at the end of BatchNorm layer~\cite{ioffe2015batch}
after each convolutional layer, since we can use the affine transformation parameters in BatchNorm to scale the channel output.
For the residual block, we only associate gates in the middle layers of each block. For the depth-wise convolution block in MobileNetV1~\cite{howard2017mobilenets},
we associate gates at the end of the second BatchNorm layer.
For the inverted residual block in MobileNetV2~\cite{sandler2018mobilenetv2}, we associate gates at the end of the first BatchNorm layer.

\section{Experiments}

\subsection{Settings}
We conduct all the experiments on CIFAR10 and ImageNet datasets. For each dataset, we allocate a separate validation set for evaluation while learning the channel gates. Specifically, we randomly select 5,000 images from the original CIFAR10 training set for validation. For ImageNet, we randomly select 50,000 images (50 images for each category) from the original training set for validation. We adopt conventional training and testing data augmentation pipelines~\cite{he2016deep}. 

When learning channel importance for the models on CIFAR10 dataset, we use Adam optimizer with an initial learning rate of 0.01 with a batch-size of 128. The balance factor $\gamma=0.5$ and total epoch is 10. All the models are expanded by $1.25\times$, and the predefined sparsity ratio $r$ equals the percentage of the pruned model's FLOPS to the full model. After searching for the pruned network architecture, we train the pruned model from scratch following the same parameter settings and training schedule in~\cite{softfilter}.

When learning channel importance for the models on ImageNet dataset, we use Adam optimizer with an initial learning rate of 0.01 and a batch-size of 100. The balance factor $\gamma=0.05$ and total epoch is 1. During training, we evaluate the model performance on the validation set multiple times. After finishing the architecture search, we train the pruned model from scratch using SGD optimizer. For MobileNets, we use cosine learning rate scheduler~\cite{loshchilov2016sgdr} with an initial learning rate of 0.05, momentum of 0.9, weight-decay of $4\times10^{-5}$. The model is trained for 300 epochs with a batch size of 256. For ResNet50 models, we follow the same hyper-parameter settings in~\cite{he2016deep}. To further improve the performance, we add label smoothing~\cite{labelsmooth} regularization in the total loss.

\begin{table}[!t]
	\centering
	\caption{Network pruning results on CIFAR10 dataset. ``Ratio'' stands for
		the percentage of the pruned FLOPS compared to the full model. Larger ratio
		stands for a more compact model. ``Baseline'' and ``Pruned'' columns stand for
		the accuracy of baseline and pruned models in percentage. ``$\Delta$ Acc''
		stands for the difference of the accuracy level between baseline and pruned model,
		and larger is better.}
	\vspace{-0.2cm}
	\scalebox{0.9}{ %
		\renewcommand\tabcolsep{2.9pt} %
		\begin{threeparttable}
			\begin{tabular}{clcccc}
				\toprule
				& \textbf{Method} & \textbf{Ratio} & \textbf{Baseline (\%)} & \textbf{Pruned (\%)} & \boldmath{}\textbf{$\Delta$ Acc (\%)}\unboldmath{} \\
				\midrule
				\multirow{6}[0]{*}{\begin{sideways}ResNet20\end{sideways}} & SFP & 40\%  & 92.20 & 90.83$\pm$0.31 & -1.37 \\
				& Rethink & 40\%  & 92.41 & 91.07$\pm$0.23 & -1.34 \\
				& Ours  & 40\%  & 91.75 & 91.14$\pm$0.32 & \textbf{-0.61} \\
				& uniform & 50\%  & 90.50 & 89.70 & -0.80 \\
				& AMC   & 50\%  & 90.50 & 90.20 & \textbf{-0.30} \\
				& Ours  & 50\%  & 91.75 & 90.55$\pm$0.14 & -1.20 \\
				\midrule
				\multirow{8}[0]{*}{\begin{sideways}ResNet56\end{sideways}} & uniform & 50\%  & 92.80 & 89.80 & -3.00 \\
				& ThiNet & 50\%  & 93.80 & 92.98 & -0.82 \\
				& CP & 50\%  & 93.80 & 92.80 & -1.00 \\
				& DCP   & 50\%  & 93.80 & 93.49 & -0.31 \\
				& AMC   & 50\%  & 92.80 & 91.90 & -0.90 \\
				& SFP & 50\%  & 93.59 & 93.35$\pm$0.31 & -0.24 \\
				& Rethink & 50\%  & 93.80 & 93.07$\pm$0.25 & -0.73 \\
				& Ours  & 50\%  & 93.23 & 93.05$\pm$0.19 & \textbf{-0.18} \\
				\midrule
				\multirow{4}[0]{*}{\begin{sideways}ResNet110\end{sideways}} & L1-norm & 40\%  & 93.53 & 93.30 & -0.23 \\
				& SFP & 40\%  & 93.68 & 93.86$\pm$0.30 & +0.18 \\
				& Rethink& 40\%  & 93.77 & 93.92$\pm$0.13 & +0.15 \\
				& Ours  & 40\%  & 93.49 & 93.69$\pm$0.28 & \textbf{+0.20} \\
				\midrule
				\multirow{6}[0]{*}{\begin{sideways}VGG16\end{sideways}} & L1-norm & 34\%  & 93.25 & 93.40 & +0.15 \\
				& NS & 51\%  & 93.99 & 93.80 & -0.19 \\
				& ThiNet & 50\%  & 93.99 & 93.85 & -0.14 \\
				& CP & 50\%  & 93.99 & 93.67 & -0.32 \\
				& DCP   & 50\%  & 93.99 & 94.16 & +0.17 \\
				& Ours  & 50\%  & 93.44 & 93.63$\pm$0.06 & \textbf{+0.19} \\
				\midrule
				\multirow{3}[0]{*}{\begin{sideways}VGG19\end{sideways}} & NS & 52\%  & 93.53 & 93.60$\pm$0.16 & +0.07 \\
				& Rethink & 52\%  & 93.53 & 93.81$\pm$0.14 & +0.28 \\
				& Ours  & 52\%  & 93.40 & 93.71$\pm$0.08 & \textbf{+0.31} \\
				\bottomrule
			\end{tabular}%
		\end{threeparttable}
	}
	\label{tab:cifar10}%
\end{table}%

\begin{table}[!t]
	\centering
	\caption{Network pruning results on ImageNet dataset. For uniform channel expansion models,
		we expand the channels of each layer with a fixed ratio $m$, denoted as ``$m\times$''. ``Baseline 1.0$\times$'' stands for
		the original full model. ``Params'' column summarizes the sizes of the total parameters of
		each pruned models.}
	\vspace{-0.2cm}
	\scalebox{0.9}{ %
		\renewcommand\tabcolsep{2.9pt} %
		\begin{threeparttable} 
			\begin{tabular}{clcccc}
				\toprule
				\multicolumn{2}{c}{\textbf{Model}} & \textbf{Params} & \textbf{Latency} & \textbf{FLOPS} & \textbf{Top-1 Acc (\%)} \\
				\midrule
				\multirow{8}[6]{*}{{\begin{sideways}MobileNet-V1\end{sideways}}} & Uniform 0.5$\times$ & 1.3M  & 20ms  & 150M  & 63.3 \\
				& Uniform 0.75$\times$ & 3.5M  & 23ms  & 325M  & 68.4 \\
				& Baseline 1.0$\times$ & 4.2M  & 30ms  & 569M  & 70.9 \\
				\cmidrule{2-6}          & NetAdapt & --    & --    & 285M  & 70.1 \\
				& AMC   & 2.4M  & 25ms  & 294M  & 70.5 \\
				\cmidrule{2-6}          & Ours 0.5$\times$ & 1.0M  & 20ms  & 150M  & \textbf{65.5} \\
				& Ours 0.75$\times$ & 1.9M  & 21ms  & 286M  & \textbf{70.7} \\
				& Ours 1.0$\times$ & 4.0M  & 23ms  & 567M  & \textbf{71.6}\\
				\midrule
				\multirow{7}[8]{*}{{\begin{sideways}MobileNet-V2\end{sideways}}} & Uniform 0.75$\times$ & 2.6M  & 39ms  & 209M  & 69.8 \\
				& Baseline 1.0$\times$ & 3.5M  & 42ms  & 300M  & 71.8 \\
				& Uniform 1.3$\times$ & 5.3M  & 43ms  & 509M  & \textbf{74.4} \\
				\cmidrule{2-6}          & AMC   & 2.3M  & 41ms  & 211M  & 70.8 \\
				\cmidrule{2-6}          & Ours 0.75$\times$ & 2.6M  & 37ms  & 210M  & \textbf{70.9} \\
				& Ours 1.0$\times$ & 3.5M  & 41ms  & 300M  & \textbf{72.1} \\
				& Ours 1.3$\times$ & 4.5M  & 42ms  & 511M  & 74.1 \\
				\midrule
				\multirow{12}[6]{*}{{\begin{sideways}ResNet50\end{sideways}}} & Uniform 0.5$\times$ & 6.8M  & 50ms  & 1.1G  & 72.1 \\
				& Uniform 0.75$\times$ & 14.7M & 61ms  & 2.3G  & 74.9 \\
				& Uniform 0.85$\times$ & 18.9M & 62ms  & 3.0G  & 75.9 \\
				& Baseline 1.0$\times$ & 25.5M & 76ms  & 4.1G  & 76.1 \\
				\cmidrule{2-6}          & ThiNet-30 & --    & --    & 1.2G  & 72.1 \\
				& ThiNet-50 & --    & --    & 2.1G  & 74.7 \\
				& ThiNet-70 & --    & --    & 2.9G  & 75.8 \\
				& SFP   & --    & --    & 2.9G  & 75.1 \\
				& CP    & --    & --    & 2.0G  & 73.3 \\
				\cmidrule{2-6}          & Ours 0.5$\times$ & 4.6M  & 44ms  & 1.0G  & \textbf{72.8} \\
				& Ours 0.75$\times$ & 9.2M  & 52ms  & 2.0G  & \textbf{75.6} \\
				& Ours 0.85$\times$ & 17.9M  & 60ms  & 3.0G  & \textbf{76.7} \\
				& Ours 1.0$\times$ & 21.5M & 67ms  & 4.1G  & \textbf{77.2} \\
				\bottomrule
			\end{tabular}%
		\end{threeparttable}
	}
	\label{tab:imagenet}%
\end{table}%

\subsection{CIFAR10 Results}
We run each experiment five times and report the ``mean $\pm$ std.'' We compare our method with other pruning methods, including naive uniform
channel number shrinkage (uniform), ThiNet~\cite{luo2017thinet},
Channel Pruning (CP)~\cite{channelpruning}, L1-norm pruning~\cite{l1norm},
Network Slimming (NS)~\cite{slimming}, Discrimination-aware Channel Pruning (DCP)~\cite{dcp},
Soft Filter Pruning (SFP)~\cite{softfilter},
rethinking the value of network pruning (Rethink)~\cite{liu2018rethinking},
and Automatic Model Compression (AMC)~\cite{he2018amc}. We compare the performance
drop of each method under the same FLOPS reduction ratio. A smaller accuracy
drop indicates a better pruning method.

Table~\ref{tab:cifar10} summarizes the results. Our method achieves
less performance drop across different model architectures compared to the
state-of-the-art methods. For large models like ResNet110 and VGGNets, our pruned
model achieves even better performance than the baseline model. Notably, our method
consistently outperforms Rethink method, which also utilizes the same budget training scheme.
This validates that our method discovers a more efficient and powerful pruned model architecture.

\subsection{ImageNet Results}

In this section, we test our method on ImageNet dataset. We mainly prune three types of
models: MobileNet-V1~\cite{howard2017mobilenets}, MobileNet-V2~\cite{sandler2018mobilenetv2}, and ResNet50~\cite{he2016deep}. We compare our method with uniform
channel expansion, ThiNet, SFP, CP, AMC, and NetAdapt~\cite{yang2018netadapt}. We report the top-1 accuracy
of each method under the same FLOPS constraint.

Table~\ref{tab:imagenet} summarizes the results. When compressing the models, our method outperforms
both uniform expansion models and other complicated pruning strategies across all three architectures.
Since our method allows the base channel expansion, we can realize the neural architecture search
by pruning the model from an enlarged supernet. Our method achieves comparable or even better
performance than the original full model design. We also measure the model CPU latency under batch size 1 on a server with two 2.40GHz Intel(R) Xeon(R) CPU E5-2680 v4. Results show that our model achieves similar or even faster model inference speed than other pruned models.  These results validate that it is both effective and scalable to prune model from a randomly initialized network directly.

\subsection{Comparison with Lottery Ticket Hypothesis}
\begin{table}[!t]
	\centering
	\caption{We compare the pruned model performance under the same pruning ratio (PR). All the models are trained for five runs on CIFAR10 dataset. ``Random'' stands for our method. ``Lottery'' stands for lottery-ticket hypothesis, which uses the original full model initialization for pruning when re-training the pruned model from scratch.}
	\vspace{-0.2cm}
	\scalebox{0.9}{ %
		\begin{tabular}{lccc}
			\toprule
			Model & PR & Random (Ours) & Lottery (Frank'19) \\
			\midrule
			ResNet20 & 40\%  & \boldmath{}\textbf{91.14$\pm$0.32}\unboldmath{} &  90.94$\pm$0.26 \\
			ResNet20 & 50\%  & \boldmath{}\textbf{90.44$\pm$0.14}\unboldmath{} & 90.34$\pm$0.36 \\
			ResNet56 & 50\%  & \boldmath{}\textbf{93.05$\pm$0.19}\unboldmath{} & 92.85$\pm$0.14 \\
			ResNet110 & 40\%  & \boldmath{}\textbf{93.69$\pm$0.28}\unboldmath{} & 93.55$\pm$0.37 \\
			VGG16 & 50\%  & \boldmath{}\textbf{93.63$\pm$0.06}\unboldmath{} & 92.95$\pm$0.22 \\
			VGG19 & 52\%  & \boldmath{}\textbf{93.71$\pm$0.08}\unboldmath{} & 93.51$\pm$0.21 \\
			\bottomrule
		\end{tabular}%
	}
	\label{tab:pretrained-lottery}%
\end{table}%

\begin{figure*}[!t]
	\centering
	\includegraphics[width=0.9\textwidth]{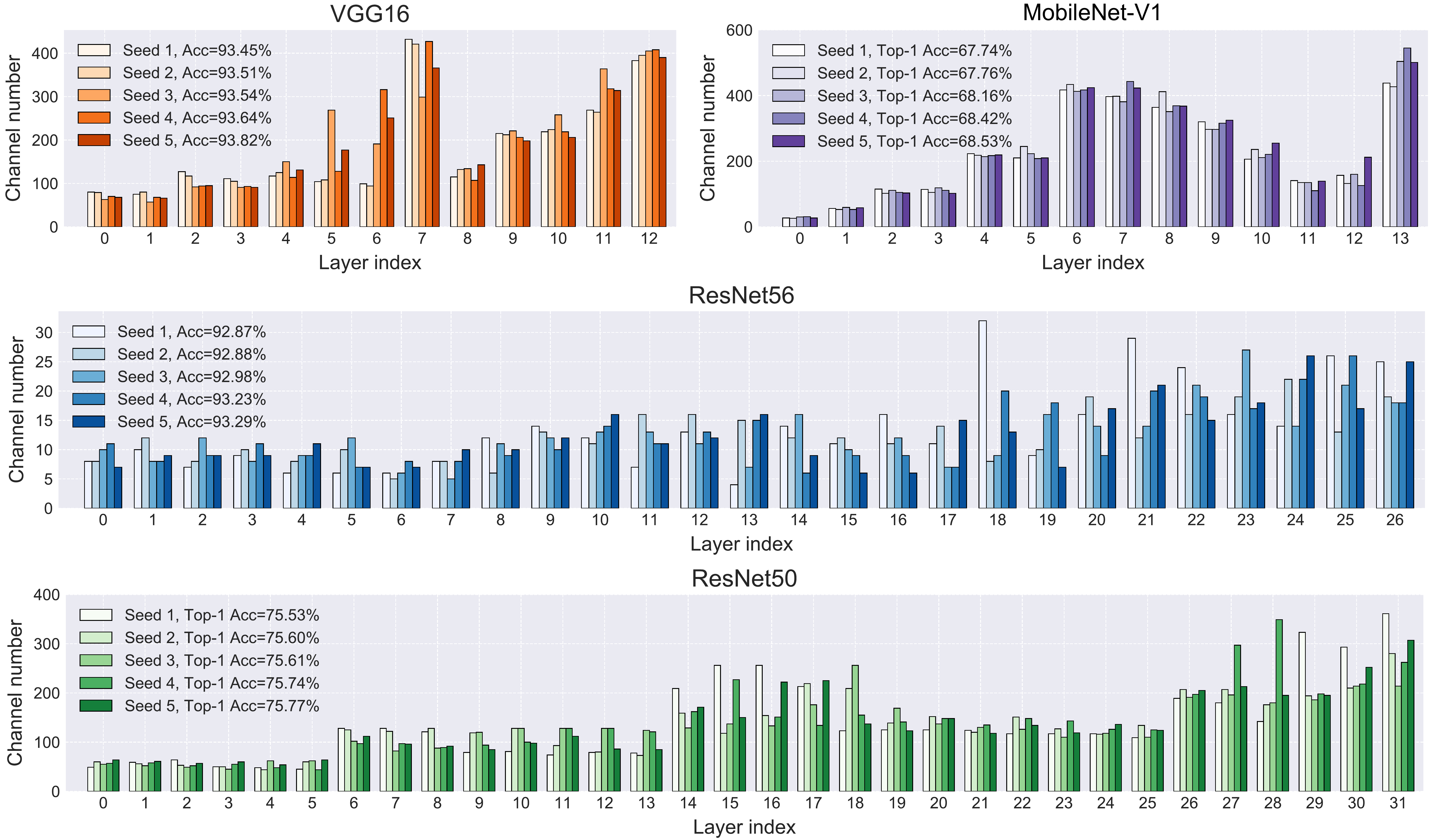}
	\caption{Visualization of channel numbers of the pruned models. For each network architecture, we learn the channel importance and prune 50\% FLOPS compared to the full model under five different random seeds. VGG16 and ResNet56 are trained on CIFAR10, and MobileNet-V1 and ResNet50 are trained on ImageNet.}
	\label{fig:channel-number}
\end{figure*}
Figure~\ref{fig:channel-number} displays the channel numbers of the pruned models on CIFAR10 and ImageNet datasets. For each network architecture, we learn the channel importance and prune 50\% FLOPS compared to the full model under five different random seeds. Though there are some apparent differences in the channel numbers of the intermediate layers, the resulting pruned model performance remains similar. This demonstrates that our method is robust and stable under different initialization methods. 

\begin{figure}[!t]
	\centering
	\includegraphics[width=0.9\columnwidth]{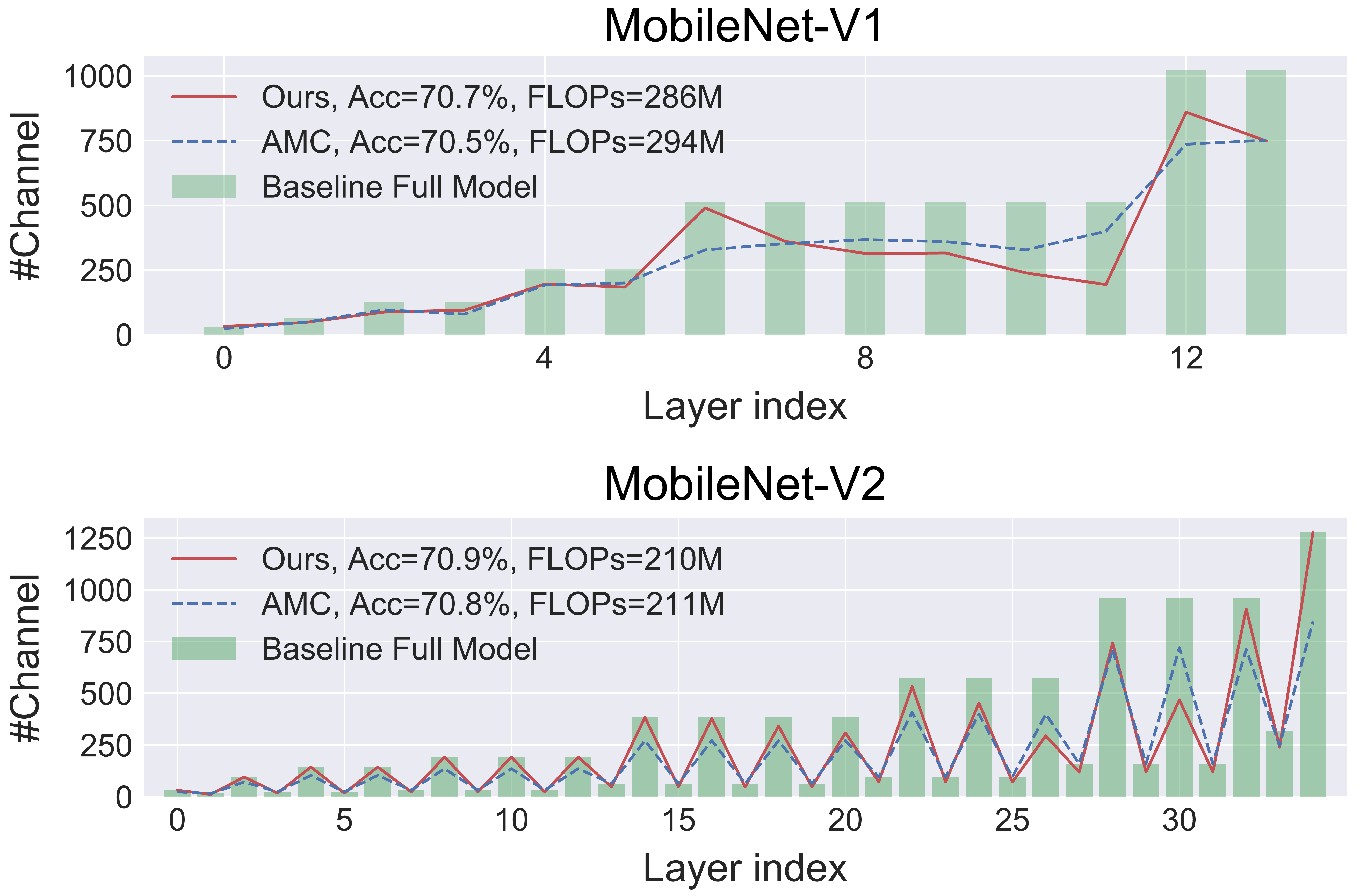}
	\caption{Pruned model structure compared with AMC. Both models are trained on the ImageNet dataset. We include the top-1 accuracy and FLOPS of each model in the legend.}
	\label{fig:compare-amc}
\end{figure}

According to the Lottery Ticket Hypothesis (LTH)~\cite{lottery}, a pruned model can only be trained to a competitive performance level if it is re-initialized to the original full model initialization weights (``winning tickets''). In our pipeline, we do not require that the pruned model has to be re-initialized to its original states for re-training the weights. Therefore, we conduct comparison experiments to testify whether LTH applies in our scenario. Table~\ref{tab:pretrained-lottery} summarizes the results. We traine all the models for five runs on CIFAR10 dataset. From the results, we conclude that our method achieves higher accuracy of the pruned models in all the cases. For Lottery Ticket Hypothesis, we do not observe the necessity of its usage. Similar phenomena are also observed in~\cite{liu2018rethinking}. There are several potential explanations. First, our method focuses on structured pruning, while LTH draws conclusions on the unstructured pruning, which can be highly sparse and irregular, and a specific initialization is necessary for successful training. Second, as pointed by~\cite{liu2018rethinking}, LTH uses Adam optimizer with small learning rate, which is different from the conventional SGD optimization scheme. Different optimization settings can substantially influence the pruned model training. In conclusion, our method is valid under the mild pruning ratio in the structured pruning situation.

\subsection{Computational Costs for Pruning}
Since our pruning pipeline does not require updating weights during structure learning, we can significantly reduce the pruned model search cost. We compare our approach to traditional Network Slimming and RL-based AMC pruning strategies. We measure all model search time on a single NVIDIA GeForce GTX TITAN Xp GPU.

When pruning ResNet56 on the CIFAR10 dataset, NS and AMC take 2.3 hours and 1.0 hours, respectively, and our pipeline only takes 0.12 hours. When pruning ResNet50 on ImageNet dataset, NS takes approximately 310 hours to complete the entire pruning process. For AMC, although the pruning phase takes about 3.1 hours, a pre-trained full model is required, which is equivalent to about 300 hours of pre-training. Our pipeline takes only 2.8 hours to obtain the pruned structure from a randomly initialized network. These results illustrate the superior pruning speed of our method.

\subsection{Visualizing Pruned Structures}

We also compare the pruned structures with those identified by AMC~\cite{he2018amc}, which utilizes a more complicated RL-based strategy to determine layer-wise pruning ratios. Figure~\ref{fig:compare-amc} summarizes the difference. On MobileNet-V1, our method intentionally reduces more channels between the eighth and eleventh layers, and increases channels in the early stage and the final two layers. The similar trend persists in the last ten layers of MobileNet-V2. This demonstrates that our method can discover more diverse and efficient structures.

\begin{figure}[htbp]
	\centering
	\includegraphics[width=\columnwidth]{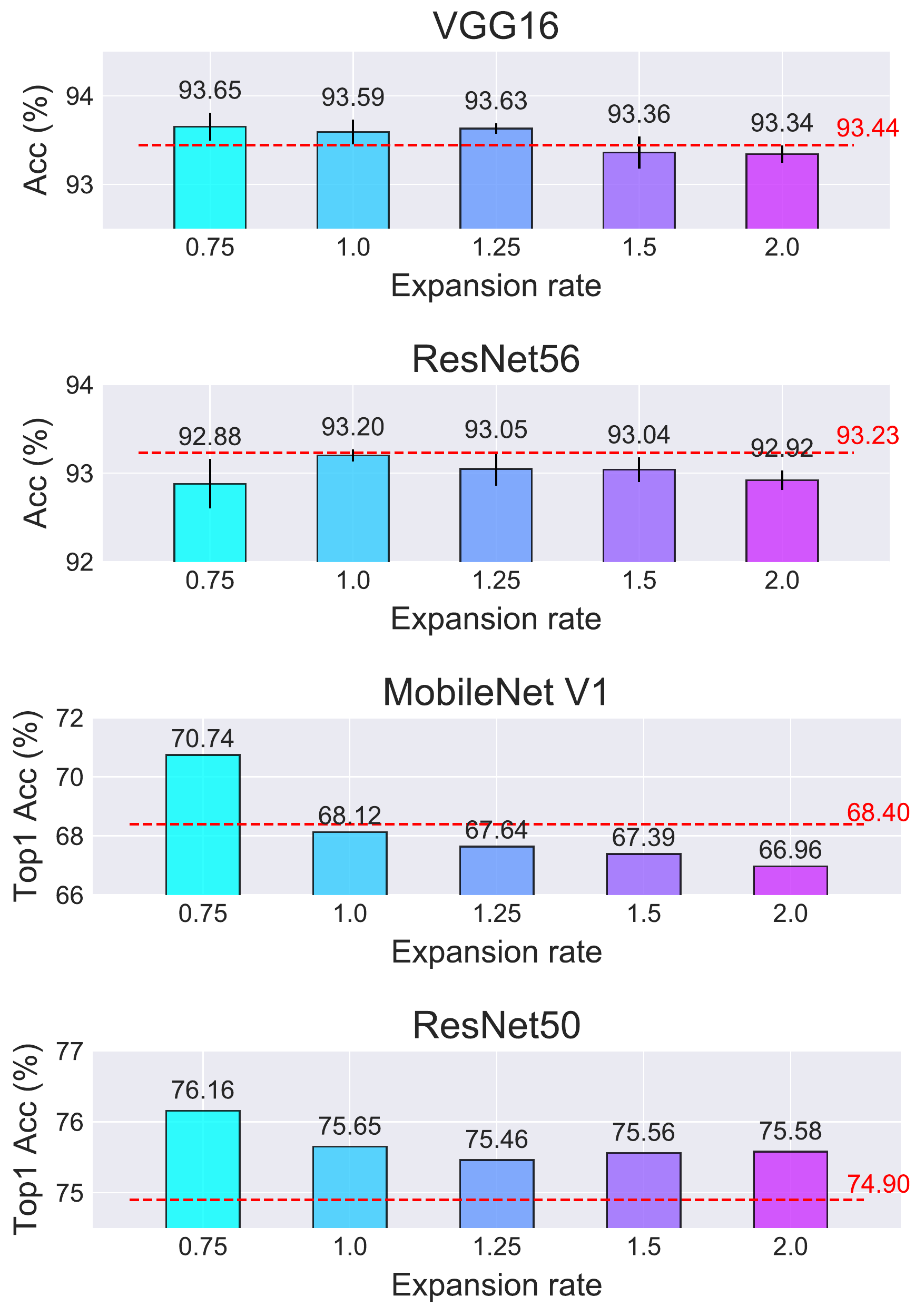}
	\caption{Effects of different expansion rate on the pruned model accuracy. Red dotted lines denote the baseline full models accuracy. VGG16 and ResNet56 models are trained on CIFAR10 dataset for five runs. MobileNet V1 and ResNet50 models are trained on ImageNet.}
	\label{fig:expansion}
\end{figure}

\begin{figure}[htbp]
	\centering
	\includegraphics[width=\columnwidth]{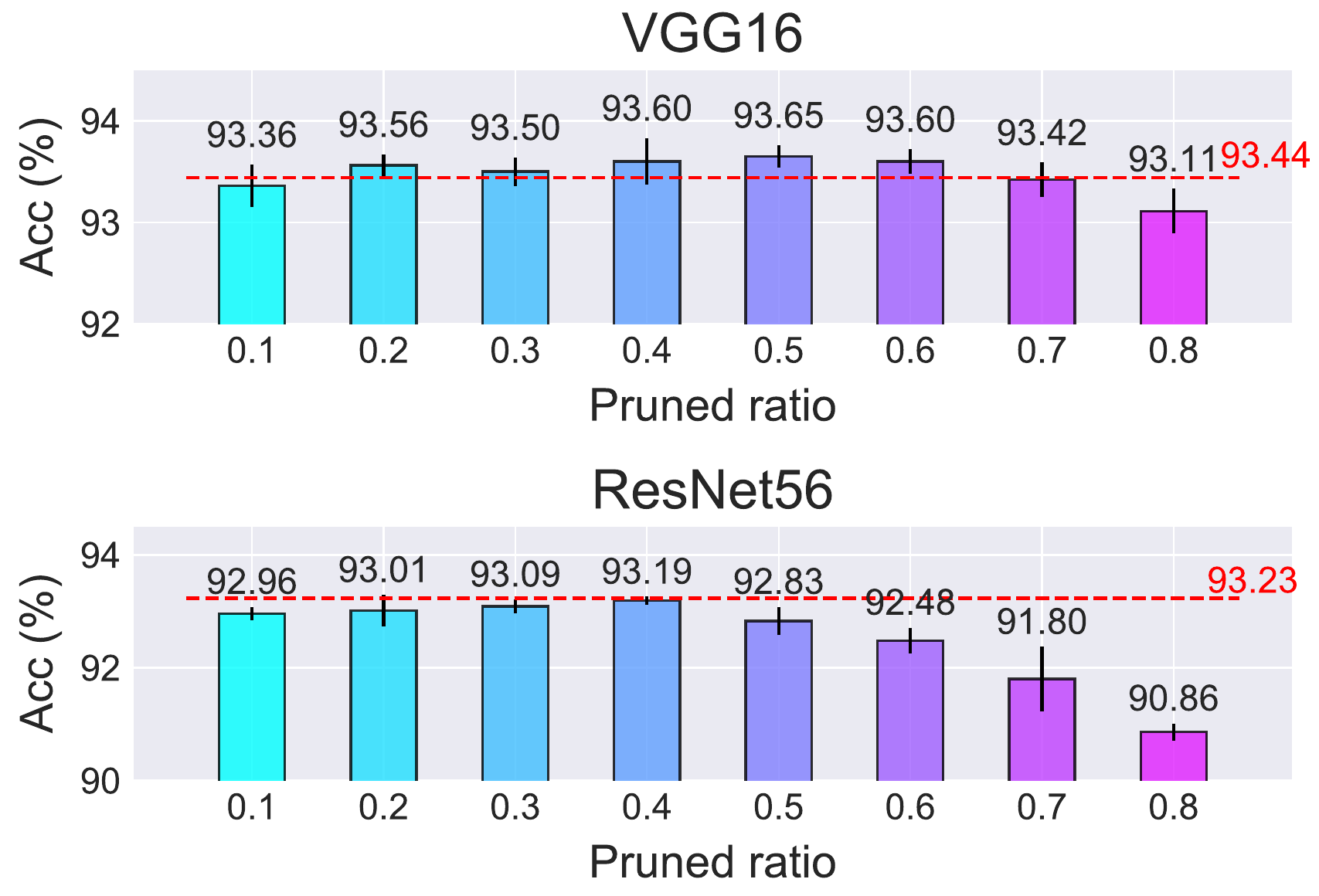}
	\caption{Effects of different pruning ratio on the model accuracy. Red dotted lines denote the baseline full models accuracy. All the models are trained on CIFAR10 dataset for five runs.}
	\label{fig:pruned-ratio}
\end{figure}

\begin{figure}[htbp]
	\centering
	\includegraphics[width=\columnwidth]{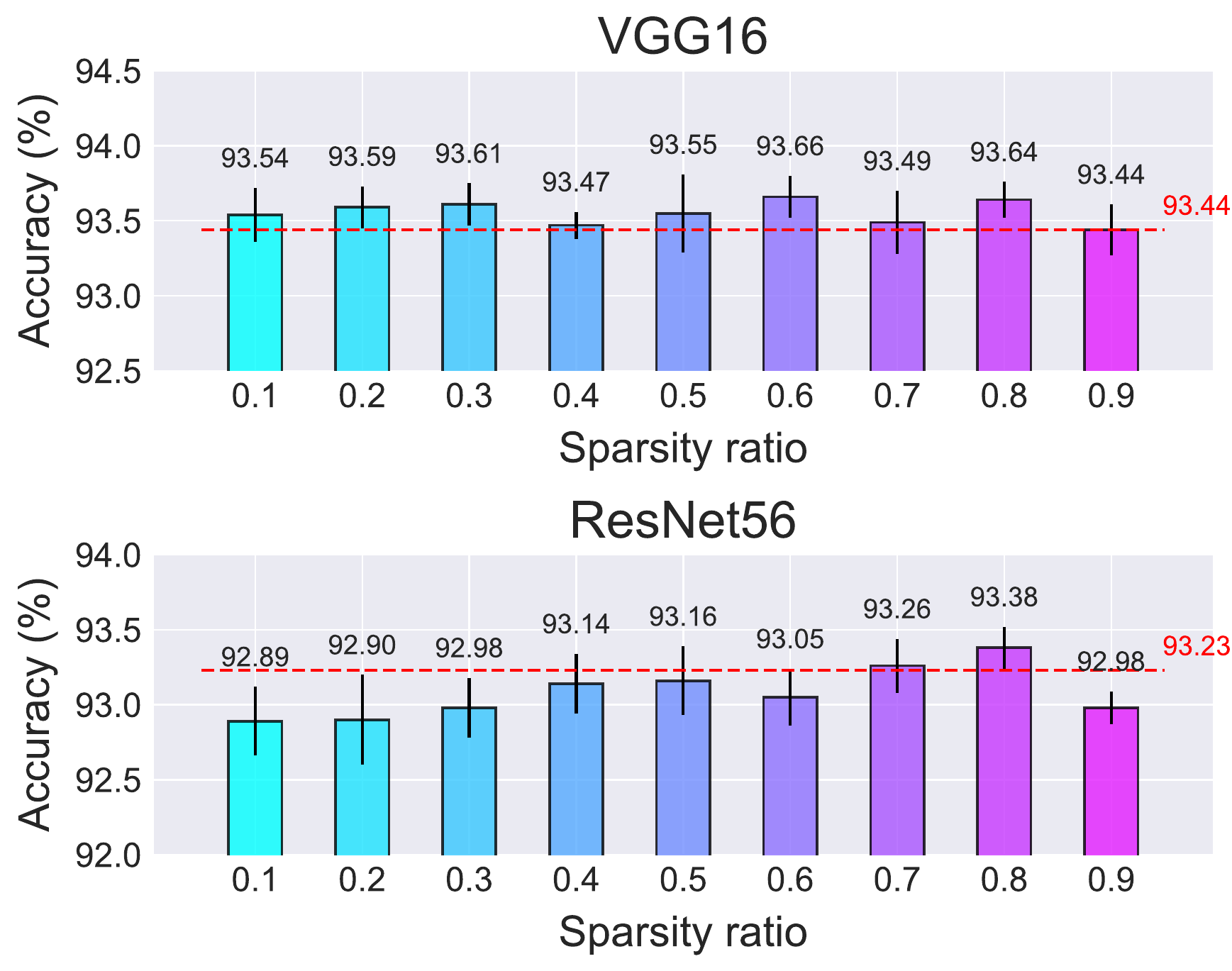}
	\caption{Effects of different sparsity ratio on the model accuarcy. Red dotted lines denote the accuracy of baseline full models.}
	\label{fig:sparsity}
\end{figure}

\section{Ablation Study}
In the following sections, we explore the performance of our method under different channel expansion rate, pruning ratio and sparsity level. 

\subsection{Channel Expansion Rate}

We have proposed to use a width multiplier to enlarge
the channels of each layer as channel expansion uniformly in the previous section. We further investigate the effect of different 
expansion rate to the final pruned model accuracy. Figure~\ref{fig:expansion} displays the results. All the pruned models are required to reduce 50\% FLOPS compared to the full models. From the figure, we find that a general trend of the influence is that when the expansion rate is too large, the pruned model performance will deteriorate. We also
surprisingly notice that using the channel shrinkage (0.75$\times$ expansion) can even achieve higher pruned model performance in some situations. This is because the preset reduced model capacity can limit the search space, which makes the pruning algorithm easier to find efficient structures. 

\subsection{Pruning Ratio}

In this section, we explore the performance of the pruned model under different pruning ratio.
Figure~\ref{fig:pruned-ratio} displays the results. For each pruned model, the channel importance is learned by setting predefined sparsity ratio $r$ as $1-\mathrm{pruning\_ratio}$. Also, all the models are trained under the same hyper-parameter settings with budget training scheme. From the figure, we conclude that our method is robust under different pruning ratio. Even under the extreme situation where a large portion of FLOPS is reduced, our method still achieves comparable prediction performance. 

\subsection{Sparsity Ratio}

In this section, we explore the effects of different sparsity ratio on the performance  of the pruned model. The predefined sparsity ratio $r$ is utilized to restrict the overall sparsity of channel importance value. Figure~\ref{fig:sparsity} summarizes the results. All the models are required to reduce 50\% FLOPS of the original full models. From the figure, we observe that the final pruned model accuracy is not very sensitive to the sparsity ratio, though a small sparsity level may have the negative impact on the performance. This demonstrates that our method is stable for a range of sparsity ratio and does not require hyper-parameter tuning.

\section{Discussion and Conclusions}
In this paper, we demonstrate that the novel pipeline of pruning from scratch is efficient and effective through extensive experiments on various models and datasets. In addition to high accuracy, pruning from scratch has the following benefits: 1) we can eliminate the cumbersome pre-training process and search the pruned structure directly on the randomly initialized weights in an extremely fast speed; 2) the pruned network structure can no longer be limited by the original network size, but can explore a larger structure space, which helps to search for better pruned model structure.

Another important observation is that pre-trained weights reduce the search space for the pruned structure. Meanwhile, we also observe that even after a short period of pre-training weights, the possible pruned structures have become stable and limited. This perhaps implies that the learning of structure may converge faster than weights. Although our pruning pipeline fixes the random initialization weights, it needs to learn the channel importance. This is equivalent to treating each weight channel as a single variable and optimizing the weighting coefficients. The pruned structure learning may become easier with reduced degree of variables.

\normalsize

{\small
\bibliographystyle{ieee}
\bibliography{reference}
}

\newpage
\appendix

\section{Effects of Pre-training on Pruning}
\begin{figure*}[htbp]
	\centering
	\includegraphics[width=\textwidth]{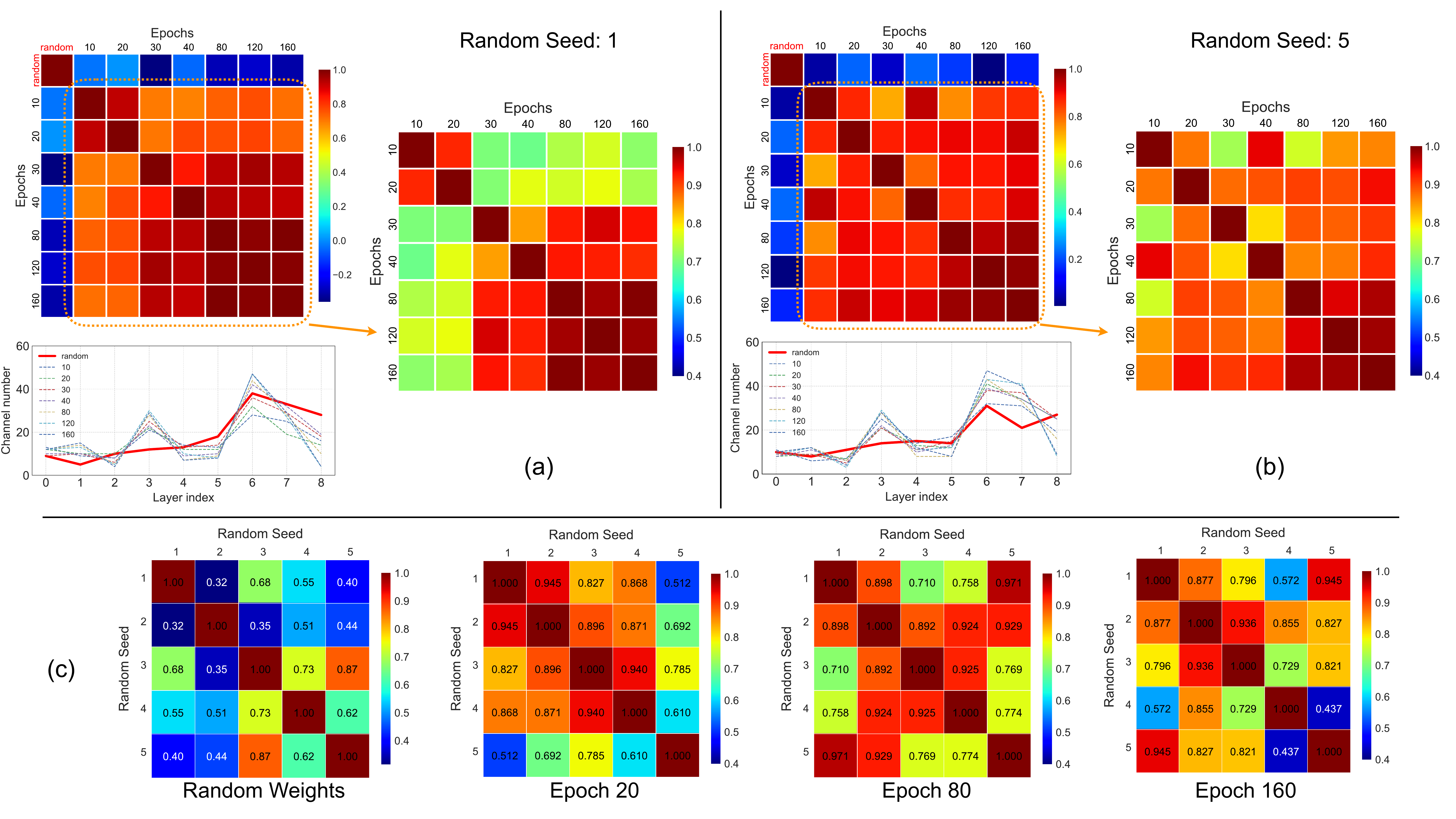}
	\caption{Exploring the effect of pre-trained weights on pruned structure  by using \textbf{ResNet20} model. }
	\label{fig:resnet20}
\end{figure*}

\begin{figure*}[htbp]
	\centering
	\includegraphics[width=\textwidth]{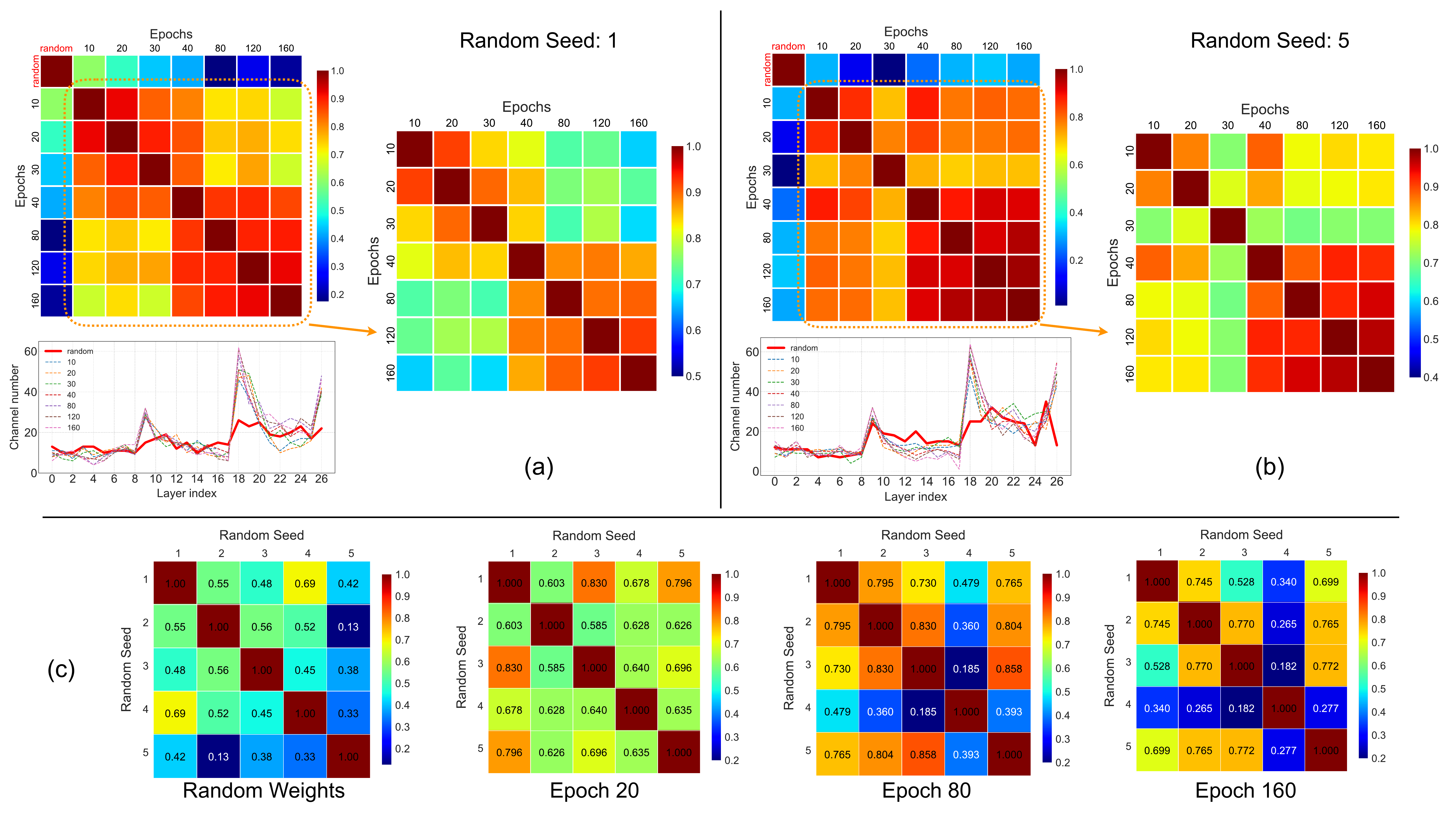}
	\caption{Exploring the effect of pre-trained weights on pruned structure by using \textbf{ResNet56} model.}
	\label{fig:resnet56}
\end{figure*}

In the main text, we explore the effects of pre-trained weights on pruned structures by visualizing the structure similarity matrices. Here we present more similar results of ResNet20 and ResNet56 models on CIFAR10 datasets.

Figure~\ref{fig:resnet20} and~\ref{fig:resnet56} show the results. All the pruned models are required to reduce 50\% FLOPS of their original models on CIFAR10 dataset. In each figure, (a) we display the correlation coefficient matrix of the pruned models directly learned from randomly initialized weights (\textbf{``random''}) and other pruned models based on different checkpoints during pre-training (\textbf{``Epochs''}) (top-left). We display the correlation coefficient matrix of pruned structures from pre-trained weights in a finer scale (right). We show the channel numbers of each layer of different pruned structures (bottom-left). Red line denotes structure from random weights; (b) similar results from the experiment with a different random seed; (c) we display correlation coefficient matrices of all the pruned structure from five different random seeds. We mark the names of initialized weights used to get pruned structure below.

For ResNet20 and ResNet50, we observe the same phenomena with those in VGG16. First, that the pruned structures learned from random weights are not similar to all the network structures obtained from pre-trained weights. Second, the pruned model structures learned directly from random weights are more diverse with various correlation coefficients. Third, the pruned structure based on the checkpoints from near epochs are more similar with high correlation coefficients in the same experiment run. 

The only difference between ResNet models with VGG16 is that the the similarities of the pruned structure based on the pre-trained weights of different random seeds are not as high as those of VGG16. This is mainly due to the fact that we only prune the layers on the residual branch in ResNet. In the case that the channel numbers of backbone layers are fixed, the number of channels of these pruned layers can have greater freedom of choice, so that they didn't converge to the same structure. However, the similarity between pruned structures based on pre-trained weights is still higher than that obtained from random weights. These results further validate our analysis in the main text.

\end{document}